%% file: main.tex
\documentclass{article}

\PassOptionsToPackage{numbers,sort&compress}{natbib}

\usepackage[preprint]{neurips_2026}

\usepackage[utf8]{inputenc} 
\usepackage[T1]{fontenc}    
\usepackage{hyperref}       
\usepackage{url}            
\usepackage{booktabs}       
\usepackage{amsfonts}       
\usepackage{nicefrac}       
\usepackage{microtype}      
\usepackage{xcolor}         
\usepackage{enumitem}
\usepackage{amsmath}

\usepackage{algorithm}
\usepackage{algpseudocode}
\usepackage{graphicx}
\usepackage{booktabs}     
\usepackage{multirow}     
\usepackage{makecell}     
\usepackage{adjustbox}    
\usepackage{setspace}     
\usepackage{xcolor,colortbl}
\usepackage{tabularx}
\usepackage{threeparttable}
\usepackage{amsthm}

\setlist[itemize]{
  leftmargin=6.0mm, 
  itemsep=5pt,
  topsep=0pt,
  parsep=0pt,
  partopsep=0pt
}

\title{CoDiffGRN: Rethinking Gene Regulatory Network Inference via the BEELINE-KGC Benchmark and Co-evolutionary Discrete Diffusion}

\author{
  Jiaze Song\\
  Peking University\\
  \texttt{jzsong25@stu.pku.edu.cn} \\
  \And
  Runhao Zhao\\
  National University of Defense Technology\\
  \texttt{runhaozhao@nudt.edu.cn} \\
  \And
  Minghao Xu\\
  Peking University\\
  \texttt{minghao.xu@stu.pku.edu.cn}\\
  \And
  Bin Cui\\
  Peking University\\
  \texttt{bin.cui@pku.edu.cn}\\
  \And
  Wentao Zhang\thanks{Corresponding author}\\
  Peking University\\
  \texttt{wentao.zhang@pku.edu.cn}
}

\newcommand{\benchmark}{BEELINE-KGC}
\newcommand{\method}{CoDiffGRN}

\begin{document}

\maketitle


\input{sections/00_abstract}
\input{sections/01b_introduction_new}
\input{sections/02_related}
\input{sections/03_benchmark}
\input{sections/04_methodology}
\input{sections/05_experiments}
\input{sections/06_conclusion}


\newpage

\bibliographystyle{unsrtnat}
\bibliography{ref}


\clearpage

\appendix

\input{appendix/B_org_results}
\input{appendix/E_cell_discrete}
\input{appendix/F_tass}
\input{appendix/G_BEELINE}

\end{document}

%% file: sections/00_abstract.tex
\begin{abstract}    \label{abs}

Inferring gene regulatory networks (GRNs) from single-cell transcriptomic data is crucial for biological discovery, yet existing approaches suffer from a fundamental misalignment with real-world needs. 
Researchers typically seek a small set of high-confidence regulatory interactions for experimental validation, often involving previously unseen genes. However, current benchmarks rely on transductive splits with global classification metrics, while prevailing models struggle to generalize under inductive settings. 
To bridge this gap, we reformulate GRN inference as an inductive, ranking-centric graph completion problem and introduce \textbf{\benchmark}, a new benchmark that incorporates an inductive gene-holdout split together with knowledge graph completion metrics to better evaluate top-ranked predictions. Building on this, we propose \textbf{\method}, the first co-evolutionary discrete diffusion framework that jointly models biologically coherent discretized gene expression states and regulatory interactions for robust inductive generalization and improved top-ranked regulatory discovery. We further introduce TF-ALL Subgraph Sampling (TASS) for scalable training. 
Extensive experiments on {\benchmark} show that {\method} establishes new state-of-the-art performance, significantly outperforming existing methods in novel regulatory discovery, and ablation studies further verify the effectiveness of our design.

\end{abstract}

%% file: sections/01b_introduction_new.tex
\section{Introduction}  \label{sec:intro}

Gene regulatory networks (GRNs) dictate cellular identity and physiological function by encoding the complex interactions between transcription factors (TFs) and target genes~\citep{wang2014protein, chronis2017cooperative, hecker2009gene, wittkopp2012cis}. While the advent of single-cell RNA sequencing (scRNA-seq) has fundamentally advanced GRN inference by resolving bulk-level signal averaging~\citep{kolodziejczyk2015technology}, the resulting data is notoriously heterogeneous and sparse~\citep{wagner2016revealing, kharchenko2014bayesian}. To tackle these challenges, recent progress has shifted from classical statistical methods to graph neural networks (GNNs), particularly graph autoencoder (GAE) architectures~\citep{chen2022graph, mao2023predicting, zhang2025inferring, hegde2026grnformer, yu2025gclink, wang2023inferring}. 
Despite these architectural advances, we argue that computational GRN inference suffers from a fundamental misalignment with practical biological discovery in two critical dimensions:

\textbf{First, there is an evaluation misalignment.} Standard benchmarks such as BEELINE~\citep{pratapa2020benchmarking} evaluate GRN inference under transductive settings, where all genes are observed during training. However, discovering regulatory mechanisms for novel modules or rare cell types inherently requires inductive generalization to unseen genes~\citep{ding2026dissecting, gilmore2024identifying}. 
Moreover, these benchmarks formalize GRN inference as a binary classification task measured by AUROC and AUPRC. In practice, however, biologists rely on AI models to prioritize a small, high-confidence set of novel interactions for costly experimental validation. Global metrics such as AUROC are easily dominated by low-ranked edges, making them poor proxies for the top-$K$ ranking quality that ultimately drives scientific discovery.

\begin{figure}[t]
\centering
    \includegraphics[width=0.95\linewidth]{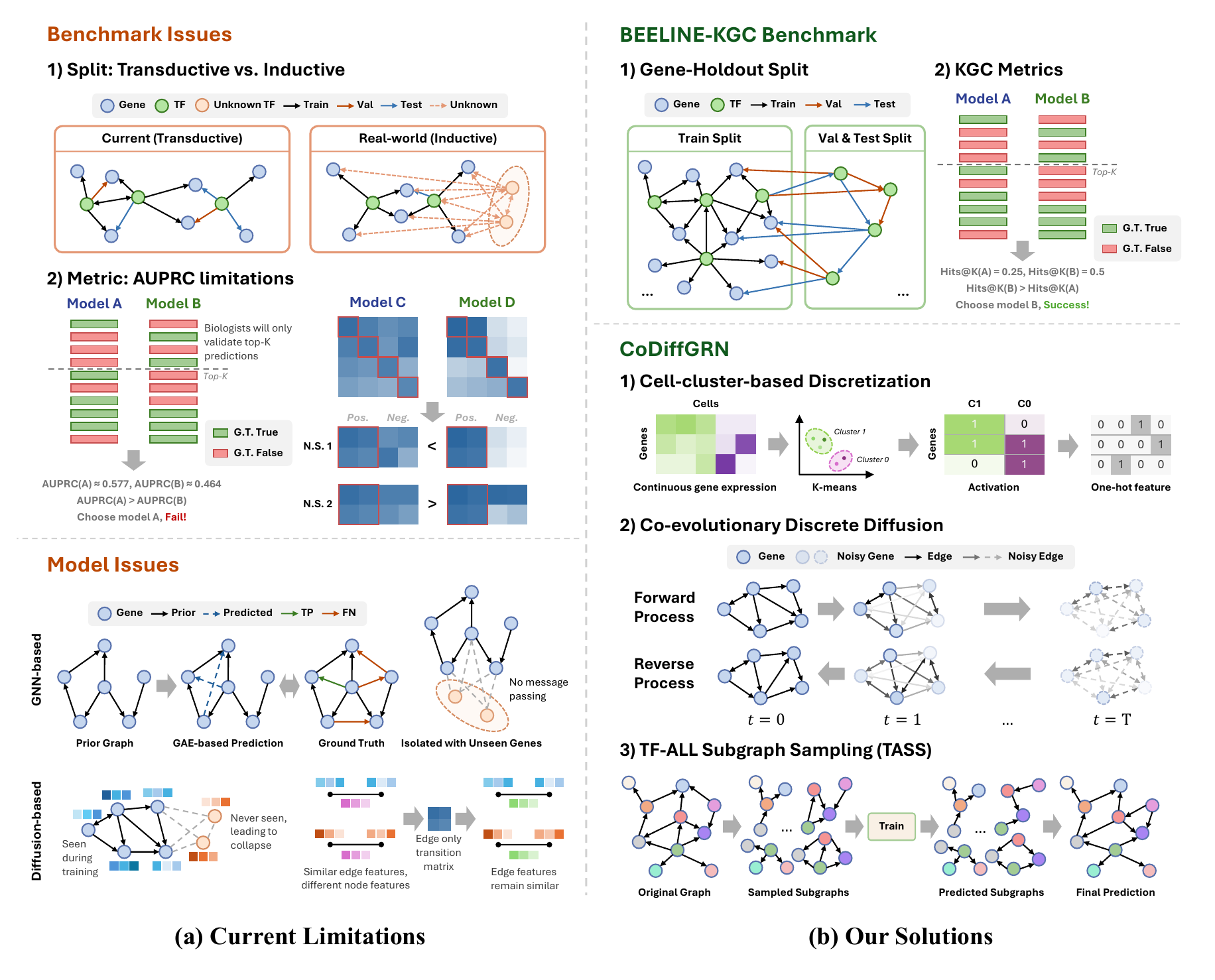}
    \caption{\textbf{Overview of current limitations and our proposed solutions.} 
    (a) Current benchmarks mismatch with realistic demand of inductive generalization and top-$K$ ranking quality, while existing methods degrade with unseen genes and fixed edge transitions.
    (b) Our {\benchmark} introduces an inductive, ranking-based evaluation protocol, and {\method} leverages cell-cluster discretization, co-evolutionary discrete diffusion with TASS. 
    \emph{Abbr.}, G.T.: Ground-truth, N.S.: Negative Sample.}
    \label{fig:intro}
\end{figure}

\textbf{Second, there exists a gene-level inductive generalization gap}. Under realistic inductive settings, existing GAE-based methods rely heavily on predefined prior graphs. When holdout genes become disconnected during inference, they lose critical contextual signals and suffer severe performance degradation (Figure~\ref{fig:intro}a). Recent diffusion-based methods partially alleviate this topological isolation, but still fail to generalize effectively to unseen genes, as the highly heterogeneous and sparse nature of scRNA-seq data prevents reliable feature correspondence and can lead to feature collapse. 

\textbf{More fundamentally, current approaches fail to capture the conditional nature of gene regulations.} 
Regulatory interactions are not isolated events, but are conditioned on the paired states of regulator and target genes. Applying a uniform transition to all edges during diffusion, regardless of the activity states of their endpoint genes, violates this fundamental principle of GRNs.

To address these limitations, we introduce a new evaluation paradigm alongside a novel generative framework. First, we propose \textbf{{\benchmark}}, a systematic extension of BEELINE that enforces an \emph{Inductive Gene-Holdout Split}. By recasting evaluation as a Knowledge Graph Completion (KGC) task~\citep{bordes2013translating}, {\benchmark} uses Hits@K and Mean Reciprocal Rank (MRR) to directly measure the top-$K$ ranking consistency of regulatory predictions. Under this rigorous protocol, existing state-of-the-art models exhibit severe generalization bottlenecks.

On the modeling side, we propose \textbf{{\method}}, the first gene-regulation co-evolutionary discrete diffusion model for GRN inference. Built upon the D3PM framework~\citep{austin2021structured}, {\method} departs from prior-dependent paradigms and is designed for robust inductive generalization. To mitigate feature collapse on unseen genes, we introduce a \textbf{Cell-cluster-based Discretization} strategy that transforms scRNA-seq profiles into biologically coherent discrete states, allowing unseen genes to inherit prior distributional structure during diffusion.

To capture the conditional nature of gene regulation, we introduce a joint node-edge diffusion process that directly models the co-evolution of discretized gene expression states and network topology at the level of diffusion dynamics. By devising a \emph{gene-state-dependent transition matrix}, edge transitions are explicitly conditioned on the states of their endpoint genes, promoting the coherent neighborhoods required for top-$K$ regulatory discovery. 
Furthermore, to enable scalable generative training under data scarcity, we introduce \textbf{TF-ALL Subgraph Sampling (TASS)}, a data augmentation strategy that preserves TF-target integrity while substantially improving scalability and generalization.

Comprehensive experimental results on {\benchmark} show that {\method} achieves state-of-the-art performance and consistently outperforms existing models. Extensive ablation studies examine the contributions of the proposed gene-regulation co-evolutionary discrete diffusion and TASS framework. These results collectively support the effectiveness of our approach in improving KGC-based ranking performance, enabling more accurate and reliable biological discovery.

%% file: sections/02_related.tex
\section{Related Work}  \label{sec:related}

\textbf{Evaluation protocols for GRN inference.} 
BEELINE~\citep{pratapa2020benchmarking} established the standard for GRN inference, primarily utilizing transductive edge-centric splits and global metrics such as AUROC and AUPRC~\citep{isewon2025benchmarking}. To improve discriminative sensitivity, negative sampling has evolved from random selection~\citep{moerman2019grnboost2, yuan2019deep} to heuristic strategies~\citep{wang2023inferring, zhang2025inferring}. 
However, these protocols exhibit a fundamental misalignment with practical biological discovery: transductive settings neglect the imperative for generalization to unseen genes, while global metrics fail to prioritize the top-$K$ candidates required for experimental validation.
In this work, we introduce \textbf{{\benchmark}} to bridge this gap by recasting evaluation as an inductive KGC task, utilizing ranking-based metrics (Hits@K, MRR) to assess top-$K$ regulatory predictions.

\textbf{Gene regulatory network inference methods.} Early GRN inference relied on statistical associations like GENIE3~\citep{huynh2010inferring} and GRNBoost2~\citep{moerman2019grnboost2}, or feature-based deep learning models such as CNNC~\citep{yuan2019deep} and DeepSEM~\citep{shu2021modeling}. Recently, graph representation learning has become the dominant paradigm, with GAE-based frameworks like GENELink~\citep{chen2022graph}, GENELink+~\citep{zhang2025inferring} and GNNLink~\citep{mao2023predicting} utilizing prior regulatory graphs, while methods like GCLink~\citep{yu2025gclink} and GRNFormer~\citep{hegde2026grnformer} introduce contrastive learning and VGAE-based frameworks. More recently, generative models such as RegDiffusion~\citep{zhu2024noise}, DigNet~\citep{wang2025diffusion}, and Planet~\citep{xu2025attention} have applied continuous and discrete diffusion processes~\citep{ho2020denoising, austin2021structured} to model the evolution nature of gene regulations.

However, existing methods face critical limitations: traditional approaches struggle with noisy single-cell data, while graph-based models often hinge on prior-graph connectivity. Furthermore, current generative frameworks struggle with feature collapse in inductive scenarios and fail to capture the essential co-evolution between gene and regulation with severe data scarcity. In this work, we propose \textbf{{\method}}, which employs cell-cluster discretization and joint discrete diffusion to model regulatory dynamics, alongside TASS specifically designed to address the challenges of data scarcity.

%% file: sections/03_benchmark.tex
\section{Rethinking for Practical Discovery: The {\benchmark} Benchmark} \label{sec:benchmark}

\input{sheets/00_benchmark}

We introduce {\benchmark}, a comprehensive benchmark designed to bridge the gap between computational GRN inference and realistic biological discovery. Built upon the BEELINE~\citep{pratapa2020benchmarking} (CC BY-NC 4.0) collection of experimental scRNA-seq datasets, {\benchmark} establishes a rigorous evaluation framework tailored to real-world laboratory constraints.

\subsection{Formalization and Problem Setting} \label{sec:benchmark:formal}
We formalize a gene regulatory network as a directed graph $\mathcal{G}=(\mathcal{V}, \mathcal{E})$, where nodes $v \in \mathcal{V}$ represent genes and edges $(u, v) \in \mathcal{E}$ denote regulatory interactions. We partition $\mathcal{V}$ into TFs ($\mathcal{V}_{TF}$) and target genes ($\mathcal{V}_{tgt}$). Within the TFs, we further distinguish \textbf{source TFs} ($\mathcal{V}_{src} \subseteq \mathcal{V}_{TF}$), defined as regulators with at least one observed outgoing edge in the ground truth.

The corresponding scRNA-seq data is represented by an expression matrix $\mathbf{X} \in \mathbb{R}^{N \times F_X}$, where $\mathbf{X}_{i,j}$ denotes the expression level of gene $v_i \in \mathcal{V}$ in cell $j \in \{1, \dots, F_X\}$. Given the expression matrix $\mathbf{X}$ and a subset of observed interactions $\mathcal{E}_{\text{obs}}$, the primary objective of GRN inference is to recover the unobserved regulatory structure by predicting potential links in $\mathcal{E} \setminus \mathcal{E}_{\text{obs}}$.

\subsection{A Unified Framework for Practical GRN Discovery} \label{sec:benchmark:framework}
Standard GRN inference benchmarks adopt a transductive setting and rely on global metrics such as AUPRC. However, this paradigm fundamentally diverges from the practical demands of real biological discovery, which requires \textbf{inductive generalization} to novel regulators and \textbf{high top-$K$ ranking quality} under limited experimental budgets.

\textbf{Pillar 1: Inductive gene-holdout split.} 
In realistic discovery scenarios, researchers predominantly focus on characterizing novel or unannotated TFs by identifying their downstream regulatory targets, necessitating an inductive setting where models must generalize to unseen regulators~\citep{subramanian2023genome, wiggers2025genome, zhang2020htftarget, zhang2021genome, niu2011diverse, bastakis2025molecular, ding2026dissecting, gilmore2024identifying}. 
To simulate this, we extract a subset of $\mathcal{V}_{src}$ as holdout TFs ($\mathcal{V}_{\mathrm{hold}}$), introducing an inductive protocol:

\begin{enumerate}[label=(\arabic*)]
    \item \textbf{Training}: The model observes only the subgraph induced by $\mathcal{V} \setminus \mathcal{V}_{hold}$. Expression profiles and all edges involving holdout TFs are withheld.
    \item \textbf{Evaluation}: The model receives the expression profiles for the full gene set $\mathcal{V}$ and predicts regulations across the entire network. Performance is evaluated solely on edges involving at least one holdout TF.
\end{enumerate}

\textbf{Pillar 2: KGC reformulation and multi-scale ranking protocol.} 
Practical biological discovery is governed by a \textbf{discrete experimental budget} $K$, where true interactions are useful only if ranked within the top-$K$ and lower ranks are effectively equivalent to being undiscovered. However, global metrics like AUPRC are insensitive to top-ranked precision, and are further distorted by the severe class imbalance in GRNs leaving many candidate pairs unevaluated under standard negative sampling.

To resolve this, we reformulate GRN inference as a \textbf{Knowledge Graph Completion (KGC)} problem. Each regulation is modeled as a structural triple $(v_i, \texttt{regulates}, v_j)$, where $v_i$ is the source TF (head) and $v_j$ is the target gene (tail). The task of finding targets for a holdout TF naturally translates to \emph{tail entity prediction}: given a query $(v_i, \texttt{regulates}, ?)$, the model must rank all candidate tails $t \in \mathcal{V}$ such that the ground-truth targets are placed as high as possible.

Aligned with this KGC formulation, we evaluate GRN inference as a resource-constrained ranking problem, adopting Hits@$K$ for top-$K$ recovery under a fixed validation budget and Mean Reciprocal Rank (MRR) for capturing overall ranking quality across heterogeneous budgets:
\begin{equation}
\text{Hits@}K = \frac{1}{|\mathcal{Q}|}\sum_{q \in \mathcal{Q}} \mathbf{1}[\text{rank}(t_q) \leq K], \qquad 
\text{MRR} = \frac{1}{|\mathcal{Q}|}\sum_{q \in \mathcal{Q}} \frac{1}{\text{rank}(t_q)},
\end{equation}
where $\mathcal{Q}$ is the set of evaluation queries targeting the holdout TFs, and $\text{rank}(t_q)$ is the rank of the ground-truth tail. 
Following standard KGC practice, we employ the \emph{filtered setting}, which excludes other known valid targets for the same query when calculating the rank. This reformulation guarantees that our benchmark directly quantifies a model's practical capacity to accelerate discovery within realistic experimental budgets.

%% file: sheets/00_benchmark.tex
\begin{table*}[t]
\centering
\caption{Statistics of the {\benchmark} benchmark. For each cell type and reference network, we report the number of cells, the number of source TFs, and the edge counts under inductive splits. Values are shown as {TFs+500 (TFs+1000)}. \emph{Abbr.}, S. TFs: Source TFs.}
\label{tab:statistics}

\setlength{\tabcolsep}{3pt}

\begin{spacing}{1.12}
\begin{adjustbox}{max width=0.9\linewidth}

\begin{tabular}{l c
ccccc ccccc}

\toprule
\multirow{2}{*}{\textbf{Cell Types}}
& \multirow{2}{*}{\textbf{\#Cells}}
& \multicolumn{5}{c}{\textbf{Specific}}
& \multicolumn{5}{c}{\textbf{Non-Specific}} \\

\cmidrule(lr){3-7}
\cmidrule(lr){8-12}

&
& \#S. TFs & \#Genes & \#Train & \#Valid & \#Test
& \#S. TFs & \#Genes & \#Train & \#Valid & \#Test \\

\midrule

hESC & 758
& 34 (34) & 815 (1260) & 3535 (5995) & 505 (544) & 505 (545)
& 283 (292) & 753 (1138) & 2546 (3531) & 447 (543) & 448 (543) \\

hHEP & 425
& 30 (31) & 874 (1331) & 7724 (12466) & 1107 (1546) & 1108 (1546)
& 322 (332) & 825 (1217) & 2993 (3490) & 568 (930) & 568 (931) \\

mDC & 383
& 20 (21) & 443 (684) & 536 (903) & 110 (145) & 110 (145)
& 250 (254) & 634 (969) & 2115 (2704) & 476 (607) & 476 (607) \\

mESC & 421
& 88 (89) & 977 (1385) & 22694 (33645) & 3459 (4575) & 3460 (4575)
& 516 (522) & 890 (1214) & 4677 (5738) & 1108 (1146) & 1108 (1146) \\

mHSC-E & 1071
& 29 (33) & 691 (1177) & 9286 (17825) & 1135 (2075) & 1136 (2075)
& 144 (147) & 442 (674) & 1002 (1370) & 211 (295) & 212 (295) \\

mHSC-GM & 889
& 22 (23) & 618 (1089) & 5796 (10901) & 784 (1617) & 784 (1617)
& 82 (88) & 297 (526) & 496 (925) & 123 (216) & 124 (217) \\

mHSC-L & 847
& 16 (16) & 525 (640) & 3354 (3887) & 522 (646) & 522 (647)
& 35 (37) & 164 (192) & 175 (217) & 52 (50) & 52 (50) \\

\midrule

\multirow{2}{*}{\textbf{Cell Types}}
& \multirow{2}{*}{\textbf{\#Cells}}
& \multicolumn{5}{c}{\textbf{STRING}}
& \multicolumn{5}{c}{\textbf{LOF/GOF}} \\

\cmidrule(lr){3-7}
\cmidrule(lr){8-12}

&
& \#S. TFs & \#Genes & \#Train & \#Valid & \#Test
& \#S. TFs & \#Genes & \#Train & \#Valid & \#Test \\

\midrule

hESC & 758
& 343 (351) & 511 (695) & 2862 (3433) & 697 (858) & 698 (858)
& -- & -- & -- & -- & -- \\

hHEP & 425
& 409 (414) & 646 (874) & 4960 (6269) & 1281 (1367) & 1282 (1367)
& -- & -- & -- & -- & -- \\

mDC & 383
& 264 (273) & 479 (664) & 3277 (4066) & 769 (916) & 769 (916)
& -- & -- & -- & -- & -- \\

mESC & 421
& 495 (499) & 638 (785) & 5099 (5830) & 1331 (1324) & 1332 (1325)
& 34 (34) & 775 (1099) & 3322 (4851) & 423 (445) & 424 (446) \\

mHSC-E & 1071
& 156 (161) & 291 (413) & 950 (1330) & 210 (248) & 211 (248)
& -- & -- & -- & -- & -- \\

mHSC-GM & 889
& 92 (100) & 201 (344) & 505 (1108) & 121 (101) & 122 (102)
& -- & -- & -- & -- & -- \\

mHSC-L & 847
& 39 (40) & 70 (81) & 89 (101) & 24 (26) & 24 (27)
& -- & -- & -- & -- & -- \\

\bottomrule
\end{tabular}

\end{adjustbox}
\end{spacing}
\vspace{-1mm}
\end{table*}

%% file: sections/04_methodology.tex
\section{Modeling Gene-Regulation Co‑evolution: The {\method} Framework}   \label{sec:method}

\subsection{Preliminaries}  \label{sec:method:preliminaries}

Diffusion models consist of a forward noising process and a reverse denoising process~\citep{ho2020denoising}. The forward process gradually corrupts data $\mathbf{x}_0$ into noise $\mathbf{x}_T$ through a Markov chain $q(\mathbf{x}_{1:T} \mid \mathbf{x}_0) = \prod_{t=1}^T q(\mathbf{x}_t \mid \mathbf{x}_{t-1})$. 
The reverse Markov process is parameterized by neural networks and progressively denoises the latent states from $\mathbf{x}_T$ to $\mathbf{x}_0$. The reverse trajectory is modeled as
\begin{equation}
    p_\theta(\mathbf{x}_{0:T}) = q(\mathbf{x}_T)\prod_{t=1}^{T} p_\theta(\mathbf{x}_{t-1}\mid \mathbf{x}_t).
\end{equation}

\textbf{Diffusion process.}
We base our formulation on discrete diffusion spaces. Following D3PM~\citep{austin2021structured}, we consider one-hot discrete random variables with $K$ categories, where $\mathbf{x}_0, \mathbf{x}_1, \dots, \mathbf{x}_t \in \{0,1\}^K$. The forward transition probabilities are encoded by matrices $Q_t \in \mathbb{R}^{K \times K}$, where $[Q_t]_{ij} = q(x_t = j \mid x_{t-1} = i)$. The transition distribution is defined as:
\begin{equation}
q(\mathbf{x}_t \mid \mathbf{x}_{t-1}) = \mathrm{Cat}(\mathbf{x}_t; \mathbf{p} = \mathbf{x}_{t-1} Q_t),
\end{equation}
where $\mathrm{Cat}(\mathbf{x}; \mathbf{p})$ denotes a categorical distribution over the one-hot vector $\mathbf{x}$ with probabilities $\mathbf{p}$. The marginal at timestep $t$ follows directly from the Markov property:
\begin{equation}
q(\mathbf{x}_t \mid \mathbf{x}_0) = \mathrm{Cat}(\mathbf{x}_t; \mathbf{p} = \mathbf{x}_0 \bar{Q}_t), \quad \text{where} \quad \bar{Q}_t = Q_1 Q_2 \cdots Q_t.
\end{equation}

\textbf{Reverse process.}
The reverse process initializes $\mathbf{x}_T \sim \pi(\mathbf{x})$ and generates $\mathbf{x}_0$ by reversing the timesteps $t = T, T-1, \dots, 0$. Following prior work~\citep{ho2020denoising, austin2021structured}, a neural network $\mathrm{nn}_\theta(\mathbf{x}_t)$ predicts $\tilde{p}_\theta(\tilde{\mathbf{x}}_0 \mid \mathbf{x}_t)$. The reverse transition is then computed as:
\begin{equation} \label{eq:rev1}
p_\theta(\mathbf{x}_{t-1} \mid \mathbf{x}_t) \propto \sum_{\tilde{\mathbf{x}}_0} q(\mathbf{x}_{t-1}, \mathbf{x}_t \mid \tilde{\mathbf{x}}_0) \tilde{p}_\theta(\tilde{\mathbf{x}}_0 \mid \mathbf{x}_t).
\end{equation}
By Bayes' theorem and the Markov property, the joint posterior admits a tractable form:
\begin{equation} \label{eq:rev2}
q(\mathbf{x}_{t-1}, \mathbf{x}_t \mid \mathbf{x}_0) = \frac{q(\mathbf{x}_{t-1} \mid \mathbf{x}_t, \mathbf{x}_0) q(\mathbf{x}_{t-1} \mid \mathbf{x}_0)}{q(\mathbf{x}_t \mid \mathbf{x}_0)} = \mathrm{Cat}\left(\mathbf{x}_{t-1}; \mathbf{p} = \frac{\mathbf{x}_t Q_t^\top \odot \mathbf{x}_0 \bar{Q}_{t-1}}{\mathbf{x}_0 \bar{Q}_t \mathbf{x}_t^\top}\right).
\end{equation}

\textbf{Choice of transition matrix and conditioning.}
Different designs of $Q_t$ induce distinct inductive biases~\citep{austin2021structured}. A simple yet effective parameterization is $Q_t = \alpha_t \mathbf{I} + (1 - \alpha_t)\mathbf{1m}^\top$, where $\mathbf{m}$ denotes the marginal distribution and $\alpha_t$ controls the noise scale. The cumulative transition admits a closed form $\bar{Q}_t = \bar \alpha_t \mathbf{I} + (1 - \bar \alpha_t)\mathbf{1m}^\top$, where $\bar \alpha_t = \prod_{\tau=1}^t \alpha_\tau$. The noise schedule is governed by a cosine function~\citep{nichol2021improved}. Furthermore, to enable conditional generation, the reverse process learns a conditional data distribution $p_\theta(\mathbf{x}_{t-1} \mid \mathbf{x}_t, \mathbf{c})$, where auxiliary condition $\mathbf{c}$ is injected into the denoiser to guide the reverse trajectory.

\begin{figure}[t]
\centering
    \includegraphics[width=1.0\linewidth]{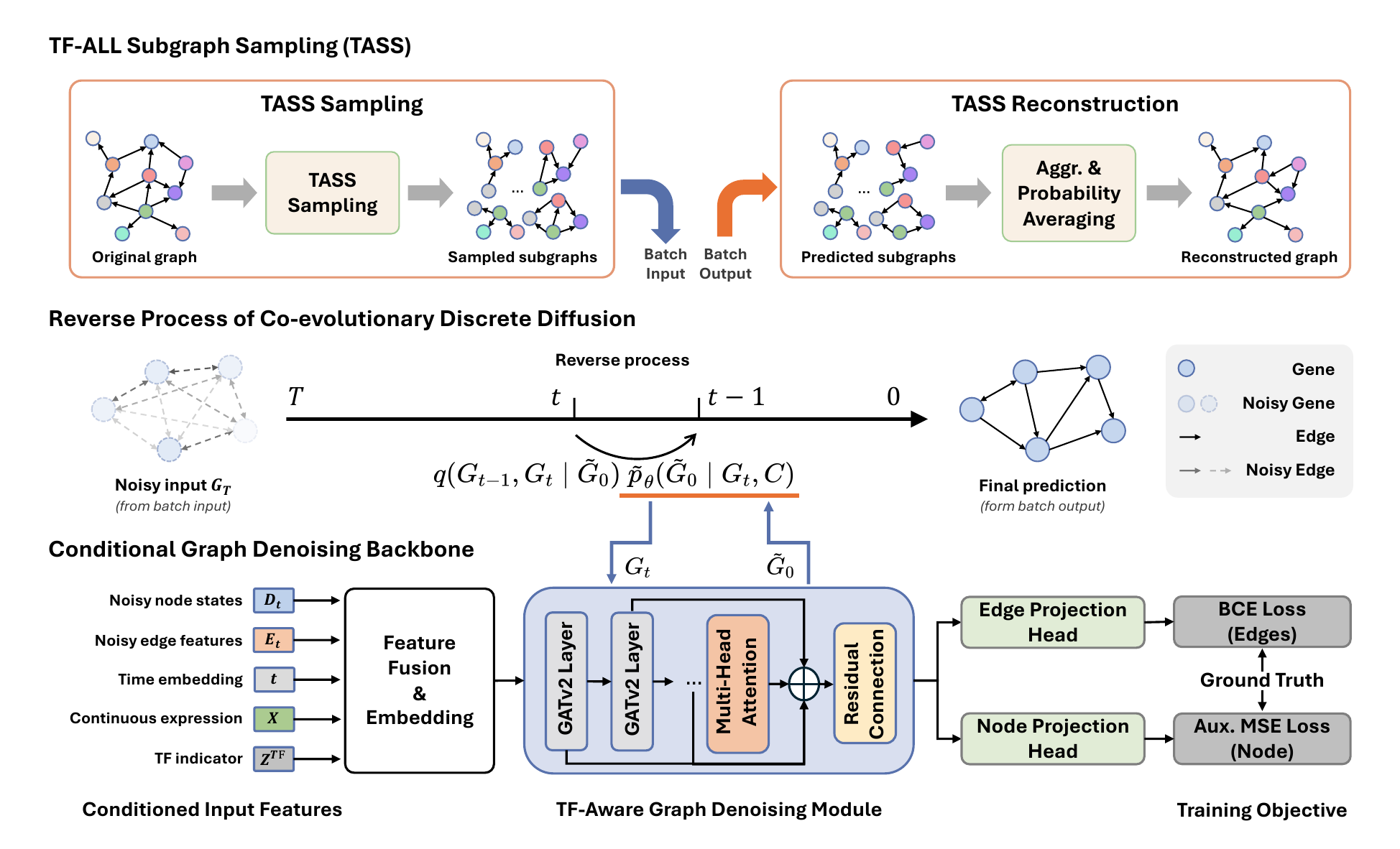}
    \vspace{-5.4mm}
    \caption{\textbf{Overview of the {\method} framework.} The framework combines TF-ALL Subgraph Sampling (TASS) with a co-evolutionary discrete diffusion model. TASS enables efficient training of a multi-conditioned TF-Aware Graph Denoising Module. At inference, predictions from multiple subgraphs are aggregated via the reverse diffusion process. \emph{Abbr.}, Aggr.: Aggregation.}
    \label{fig:method}
\end{figure}
\subsection{Joint Discrete Diffusion on Gene Regulatory Network}    \label{sec:method:joint}

As illustrated in Figure~\ref{fig:method}, we propose a joint discrete diffusion framework for GRNs. In contrast to prior works that decouple gene features from edge diffusion~\citep{zhu2024noise, wang2025diffusion}, our framework unifies discretized gene expression and regulatory interactions into a single co-evolutionary state space.

\textbf{Cell-cluster discretization.} 
To mitigate feature collapse on unseen genes, we transform continuous scRNA-seq profiles into biologically coherent discrete states. Prior studies show that discretized gene activation patterns capture stable functional structures across diverse biological settings~\citep{grun2015single, bouland2021differential}. We cluster cells into $k$ groups and binarize gene expression within each cluster via thresholding, producing a $k$-bit activation pattern per gene. Each binary pattern is mapped to a categorical variable and represented as a one-hot vector in $\mathbb{R}^{F_D}$, where $F_D = 2^k$. 

\textbf{Joint modeling.} 
We follow a joint diffusion formulation~\citep{liu2024graph} to unify gene expression and regulatory edges. Edges are represented as $\mathbf{E} \in \mathbb{R}^{N \times N \times 2}$ (indicating absence/presence), and node features as $\mathbf{D} \in \mathbb{R}^{N \times F_D}$. The global graph token for each node $i$ is constructed via concatenation: $\mathbf{G}[i] = \big[ \mathbf{d}_i \;\Vert\; \mathbf{e}_{i1} \;\Vert\; \cdots \;\Vert\; \mathbf{e}_{iN} \big]$, where $\mathbf{d}_i \in \mathbb{R}^{F_D}$ denotes the discretized node features, and $\mathbf{e}_{ij} \in \mathbb{R}^2$ is the one-hot encoding of the edge from node $i$ to node $j$.

\textbf{Design of the transition matrix.} 
We formulate a joint transition matrix $Q_G \in \mathbb{R}^{F_G \times F_G}$, constructed from four component transition probability matrices: $Q_D$ (gene $\to$ gene), $Q_E$ (regulation $\to$ regulation), $Q_{ED}$ (regulation $\to$ gene), and $Q_{DE}$ (gene $\to$ regulation):
\begin{equation}
Q_G = \begin{bmatrix}
Q_D & \mathbf{1}_N^\top \otimes Q_{DE} \\
\mathbf{1}_N \otimes Q_{ED} & \mathbf{1}_{N \times N} \otimes Q_E
\end{bmatrix},
\end{equation}
where $\otimes$ denotes the Kronecker product, and $\mathbf{1}_N$, $\mathbf{1}_{N \times N}$ are all-ones vectors and matrices of appropriate dimensions. Following~\citet{vignac2022digress}, $Q_D$ and $Q_E$ are derived from the marginal distributions $\mathbf{m}_D$ and $\mathbf{m}_E$ computed from the training data. For the cross-dependencies, $\mathbf{m}_{ED}$ and $\mathbf{m}_{DE}$ capture the conditional co-occurrence probabilities between gene expression types and regulation types. This explicitly ensures that edge transitions are biologically conditioned on the states of their endpoint genes.

\textbf{Forcing true D technique.} 
During training, the reverse process approximates $p_\theta(G_{t-1} \mid G_t) \propto \sum_{\tilde{G}_0} q(G_{t-1}, G_t \mid \tilde{G}_0) \tilde{p}_\theta(\tilde{G}_0 \mid G_t)$. To mitigate error accumulation and sharpen the learning signal for unknown edge structures, we apply a teacher-forcing-style technique that replaces the predicted node component $\tilde{\mathbf{D}}_0$ within the generated $\tilde{G}_0$ with the ground-truth $\mathbf{D}_0$.

\subsection{Conditional Graph Denoising Architecture}

\textbf{Conditioning mechanism.}
We incorporate multiple sources of conditioning information into the denoiser to guide the reverse process. Combining sinusoidal timestep embeddings $\mathbf{t}_i$, continuous gene expression features $\mathbf{x}_i$, noisy discrete node states $\mathbf{d}_{t,i}$, flattened incoming/outgoing edge features $\mathbf{e}_{t,i:}$, and a boolean TF indicator $\mathbf{z}_i^{\text{TF}}$, the conditioned input feature for node $i$ is constructed as:
\begin{equation}
\mathbf{h}_i^{(0)} = \big[ \mathbf{t}_i \;\Vert\; \mathbf{x}_i \;\Vert\; \mathbf{d}_{t,i} \;\Vert\; \mathbf{e}_{t,i:} \;\Vert\; \mathbf{z}_i^{\text{TF}} \big].
\end{equation}

\textbf{Denoiser backbone.}
We parameterize the reverse process using a TF-aware graph denoiser built upon GATv2~\citep{brody2021attentive}. Each layer updates node representations dynamically:
\begin{equation}
\mathbf{h}_i^{(l+1)} = \sigma\left( \sum_{j \in \mathcal{N}(i)} \alpha_{ij}^{(l)} \mathbf{W}^{(l)} \mathbf{h}_j^{(l)} \right),
\end{equation}
where the attention coefficients $\alpha_{ij}^{(l)}$ are computed via:
\begin{equation}
\alpha_{ij}^{(l)} = \mathrm{softmax}_j \left( \mathbf{a}^\top \, \mathrm{LeakyReLU}\big( \mathbf{W}^{(l)}[\mathbf{h}_i^{(l)} \Vert \mathbf{h}_j^{(l)}] \big) \right).
\end{equation}
The node representations are then projected into TF-specific ($\mathbf{h}^{\text{tf}}_i$) and target-specific ($\mathbf{h}^{\text{tg}}_j$) embedding spaces. A score head yields a scalar compatibility score $s_{ij} = f_\mathrm{score}([\mathbf{h}^{\text{tf}}_i \Vert \mathbf{h}^{\text{tg}}_j])$. Binary edge logits are formed as $[-s_{ij},\, s_{ij}]$ and added residually to the noisy edge input $\mathbf{E}_t$. Node logits are obtained via a linear projection of averaged embeddings with a residual connection to $\mathbf{D}_t$.

\textbf{Training objective.} 
The denoiser is trained using a weighted combination of discrete edge reconstruction loss ($\mathcal{L}_{\text{edge}}$) and a node-level auxiliary regression objective ($\mathcal{L}_{\text{node}}$). Let $\hat{{e}}_{i,j}$ denote the predicted probability of edge existence and ${e}_{i,j} \in \{0,1\}$ the ground-truth. The total loss is defined as:
\begin{equation}
\mathcal{L} = - \sum_{i,j} \left( e_{i,j} \log \hat{e}_{i,j} + (1 - e_{i,j}) \log (1 - \hat{e}_{i,j}) \right) + \lambda \frac{1}{N} \sum_{i=1}^{N} \left\| \hat{\mathbf{X}}_{0,i} - \mathbf{X}_i \right\|_2^2,
\end{equation}
where the auxiliary MSE loss preserves continuous biological signals within the discretized space, and $\lambda \ll 1$ controls its strength.

\subsection{TF-ALL Subgraph Sampling (TASS)}    \label{sec:method:tass}

Diffusion models typically require large training datasets, but GRN inference is inherently data-scarce since each dataset constitutes a single large graph. To address this, we propose TF-ALL Subgraph Sampling (TASS), which constructs a training set by sampling subgraphs from the input graph (Figure~\ref{fig:intro}(c)). The model is trained on these subgraphs and, at inference time, repeatedly samples subgraphs, predicts edges, and aggregates results to reconstruct the full graph. This expands effective supervision by exposing diverse gene–topology patterns, improving generalization under limited data.

\textbf{Sampling.} 
As shown in Figure~\ref{fig:method}, we adopt node-based sampling due to the lack of reliable edge priors, following uniform node sampling~\citep{limnios2023sagess}. Exploiting the bipartite-like structure of GRNs (TFs $\to$ targets), each subgraph of size $k$ is formed by sampling a fraction $t$ (TF-to-ALL ratio) of nodes from the TF set, and the remainder from all genes, with resampling for TF duplicates. Let $p=k/n$. Following theoretical bounds for graph recovery~\citep{limnios2023sagess}, we set the number of subgraphs to $m=\lceil p^{-2}\log n \log(1/\delta)\rceil$. In GRNs, this bipartite sampling strategy inherently yields higher edge coverage due to the sparse, hierarchical nature of regulatory interactions.

\textbf{Reconstruction.} 
At inference, we apply the same sampling strategy, generate predictions across subgraphs, and map them back to the global edge indices. Outputs are converted to probabilities via softmax, and overlapping predictions are averaged to form the final consensus. Uncovered edges are assigned a default baseline prior.

%% file: sections/05_experiments.tex
\section{Experiments}   \label{sec:exp}

\subsection{Experimental Setup} \label{sec:exp:setup}

\textbf{Evaluation protocols.} 
We primarily evaluate performance using Hits@K and MRR on Inductive Gene-Holdout Split introduced in Section~\ref{sec:benchmark}. To reflect practical experimental settings, we report Hits@10 and Hits@50, corresponding to low- and high-throughput validation regimes.

\textbf{Baseline models.} 
We compare against a comprehensive set of baselines, including two traditional methods (GENIE3~\citep{huynh2010inferring}, GRNBoost2~\citep{moerman2019grnboost2}), two feature-based deep learning models (CNNC~\citep{yuan2019deep}, DeepSEM~\citep{shu2021modeling}), four graph-based models (GNNLink~\citep{mao2023predicting}, GENELink+~\citep{zhang2025inferring}, GRNFormer~\citep{hegde2026grnformer}, GCLink~\citep{yu2025gclink}), and two diffusion-based methods (RegDiffusion~\citep{zhu2024noise}, DigNet~\citep{wang2025diffusion}).

\textbf{Model setups.} 
For fair comparison, all baselines are implemented following their original configurations. 
Our model adopts the TF-Aware Graph Denoiser, built upon a multi-layer GATv2 architecture with three attention heads per layer and progressively reduced hidden dimensions (128, 64, 64, 32), producing a 16-dimensional output embedding. 
We use a LeakyReLU with negative slope 0.2 and concatenate multi-head outputs. 
Score head $f_\mathrm{score}$ is implemented as a two-layer MLP.

\textbf{Training setups.} 
All baselines are trained according to their original protocols. Our model is optimized using AdamW with a learning rate of $1 \times 10^{-4}$, weight decay $1 \times 10^{-3}$, batch size 64, and trained for 500 epochs. We use a discrete state space with $k=4$ clusters (yielding 16 classes) and 500 diffusion steps. Subgraphs are sampled with 100 nodes each. 
All experiments are conducted on a local server with 100 CPU cores and 4 NVIDIA RTX 4090 GPUs (24 GB), and are run over three random seeds (0, 1, and 2), with both mean and standard deviation reported.

\input{sheets/01_main}

\subsection{Results}    \label{sec:exp:results}

We report the performance of representative models on the Specific network under the proposed {\benchmark} in Table~\ref{tab:benchmark}. 
Based on these results, we highlight the following key findings:

\begin{itemize}
	\item \textbf{The proposed {\benchmark} benchmark is substantially more challenging than existing protocols.} 
    Traditional and feature-based methods largely collapse under inductive gene holdout, yielding near-zero Hits@10 and MRR. Even several GNN-based models degrade substantially, indicating that {\benchmark} reveals inductive generalization failures obscured by transductive benchmarks and global metrics.
  
	\item \textbf{GNN-based models show limited inductive generalization.} 
	While GENELink+ remains competitive, most GNN methods suffer from low performance. Their heavy reliance on predefined prior graphs leads to disconnected components for unseen genes, causing error accumulation during message passing and widening the gap against diffusion-based approaches.

	\item \textbf{Diffusion-based models offer partial robustness yet remain insufficient.} According to Table~\ref{tab:benchmark}, diffusion-based methods generally outperform GNN baselines and match GENELink+ in some cases, but still fall behind {\method}. This suggests that existing diffusion models alleviate reliance on explicit graph structure, yet remain vulnerable to feature collapse on unseen genes. Moreover, their single-graph optimization paradigm appears suboptimal for diffusion.

	\item \textbf{{\method} achieves consistent SOTA performance.} 
	{\method} ranks first across all cell types and metrics in both TFs+500 and TFs+1000 settings, with relative gains up to +184.8\% (and +24.5\% on average) over the strongest baselines on the most competitive Specific network. These results validate the effectiveness of our co-evolutionary discrete diffusion framework with gene-state-dependent transitions and the TASS strategy for addressing inductive GRN inference and top-$K$ ranking quality.

\end{itemize}

\begin{figure}[t]
\centering
    \includegraphics[width=1.0\linewidth]{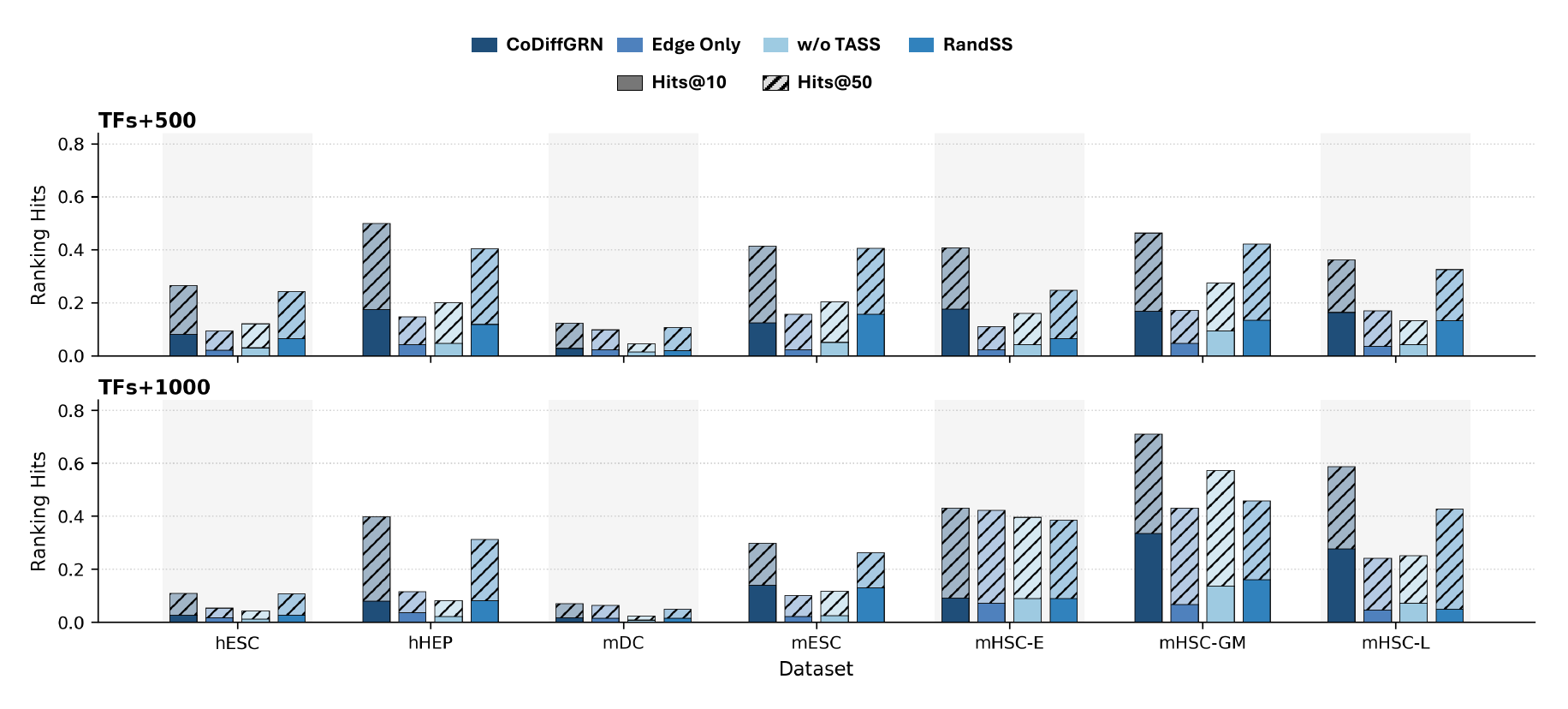}
    \vspace{-5.4mm}
    \caption{Ablation study results on {\benchmark} (\textbf{Specific} network). 
    We report \emph{mean} performance of Hits@10 and Hits@50 over runs, under the TFs+500 (top) and TFs+1000 (bottom) settings. \emph{Abbr.}, w/o: without.}
    \label{fig:ablation}
\end{figure}

\subsection{Ablation Study} \label{sec:exp:ablation}

To verify the effectiveness of our designs, we conduct extensive ablation experiments on {\benchmark}, focusing on the joint diffusion mechanism and the TASS augmentation strategy. The variants are: (i) \textbf{Edge Only}: a variant that only using edge diffusion; (ii) \textbf{w/o TASS}: the model trained on the original graph without TASS; and (iii) \textbf{RandSS}: replacing TASS with uniform random subgraph sampling.

\textbf{Effect of joint modeling.} 
As shown in Figure~\ref{fig:ablation}, removing joint modeling (Edge Only) consistently degrades performance across datasets and metrics. This indicates that decoupling gene expression states from regulatory edges limits the model’s ability to capture conditional regulatory interactions and reduces top-$K$ ranking quality, especially when generalizing to unseen genes in inductive settings.

\textbf{Effect of subgraph sampling.} 
Removing TASS leads to a substantial performance drop across all experimental settings. In contrast, uniform random sampling (RandSS) partially recovers the performance and achieves superior results in a few isolated cases, but remains consistently less effective overall. These results highlight the importance of informative subgraph construction and demonstrate that the task-aware sampling strategy of TASS provides more reliable improvements than generic random sampling under inductive and data-scarce GRN inference settings.

%% file: sheets/01_main.tex
\begin{table}[t]
\centering
\caption{Benchmark results on {\benchmark} (\textbf{Specific} network). We report \emph{mean} over runs. The best, second-best, and third-best are denoted by \textbf{bold}, \underline{underline}, and \textit{italic}, respectively. $\Delta$ indicates change relative to the baseline: {\color{green!50!black}green} up, {\color{red!70!black}red} down. Leading zeros are omitted. \emph{Abbr.}, Trad.: Traditional methods, DL.: Feature-based deep learning models, GNN.: GNN-based models, Diff.: Diffusion-based models.}
\label{tab:benchmark}

\setlength{\tabcolsep}{3pt}

\begin{spacing}{1.2}
\begin{adjustbox}{max width=0.95\linewidth}
\begin{tabular}{clccccccccccccccccccccc}

\toprule
\multicolumn{2}{c}{\multirow{2}{*}{\textbf{Model}}}
& \multicolumn{3}{c}{\bf{hESC}} 
& \multicolumn{3}{c}{\bf{hHEP}} 
& \multicolumn{3}{c}{\bf{mDC}} 
& \multicolumn{3}{c}{\bf{mESC}} 
& \multicolumn{3}{c}{\bf{mHSC-E}} 
& \multicolumn{3}{c}{\bf{mHSC-GM}} 
& \multicolumn{3}{c}{\bf{mHSC-L}} \\

\cmidrule(lr){3-5} \cmidrule(lr){6-8} \cmidrule(lr){9-11} \cmidrule(lr){12-14} 
\cmidrule(lr){15-17} \cmidrule(lr){18-20} \cmidrule(lr){21-23}

& & H@10 & H@50 & MRR & H@10 & H@50 & MRR & H@10 & H@50 & MRR 
& H@10 & H@50 & MRR & H@10 & H@50 & MRR 
& H@10 & H@50 & MRR & H@10 & H@50 & MRR \\

\midrule
\rowcolor{gray!8}
\multicolumn{23}{c}{\bf{Setting: TFs+500}} \\
\midrule
\multirow{2}{*}{\rotatebox{90}{\small \textsc{Trad.}}}
& GENIE3 & .000 & .002 & .003 & .001 & .004 & .007 & .000 & .009 & .003 & .004 & .013 & .007 & .001 & .005 & .004 & .003 & .006 & .005 & .004 & .004 & .005 \\
& GRNBoost2 & .000 & .002 & .003 & .000 & .004 & .006 & .000 & .009 & .003 & .001 & .013 & .006 & .000 & .005 & .004 & .003 & .006 & .006 & .000 & .004 & .004 \\
\midrule
\multirow{2}{*}{\rotatebox{90}{\small \textsc{DL.}}}
& CNNC & .006 & .006 & .005 & .009 & .009 & .013 & .009 & .009 & .004 & .026 & .036 & .019 & .000 & .005 & .004 & .000 & .000 & .005 & .000 & .000 & .004 \\
& DeepSEM & .000 & .004 & .003 & .001 & .002 & .007 & .000 & .000 & .003 & .005 & .017 & .007 & .000 & .002 & .004 & .000 & .003 & .005 & .000 & .000 & .004 \\
\midrule
\multirow{4}{*}{\rotatebox{90}{\small \textsc{GNN.}}}
& GNNLink & .026 & .110 & .015 & .059 & .200 & .045 & .012 & .037 & .004 & .037 & .157 & .020 & .049 & .141 & .032 & .044 & .119 & .037 & .055 & .152 & .024 \\
& GENELink+ & \textit{.052} & \textit{.218} & \underline{.030} & \underline{.141} & \underline{.385} & \underline{.100} & \underline{.025} & \underline{.077} & \underline{.009} & \underline{.082} & \underline{.343} & \underline{.050} & .124 & .355 & .073 & \underline{.115} & \underline{.290} & \underline{.073} & \textit{.122} & \textit{.304} & \textit{.067} \\
& GRNFormer & .036 & .150 & .020 & .093 & .248 & .072 & .017 & .047 & .006 & \textit{.061} & \textit{.250} & .030 & .076 & .189 & .052 & .069 & .172 & .044 & .077 & .216 & .038 \\
& GCLink & .033 & .131 & .016 & .079 & .224 & .060 & .014 & .040 & .005 & .046 & .202 & .024 & .066 & .180 & .043 & .057 & .148 & .040 & .064 & .170 & .030 \\
\midrule
\multirow{2}{*}{\rotatebox{90}{\small \textsc{Diff.}}}
& RegDiffusion & .033 & .151 & .021 & .077 & .230 & .058 & \textit{.023} & \textit{.071} & \textit{.009} & .051 & .195 & .033 & \textit{.133} & \underline{.362} & \textit{.074} & \textit{.094} & \textit{.249} & \textit{.065} & .066 & .162 & .036 \\
& DigNet & \underline{.063} & \underline{.242} & \textit{.029} & \textit{.138} & \textit{.355} & \textit{.097} & .019 & .055 & .006 & .051 & .236 & \textit{.035} & \underline{.168} & \textit{.361} & \underline{.088} & .079 & .216 & .056 & \underline{.139} & \underline{.314} & \underline{.085} \\
\midrule
\rowcolor{gray!18}
& {\method} & \textbf{.079} & \textbf{.264} & \textbf{.034} & \textbf{.173} & \textbf{.499} & \textbf{.126} & \textbf{.028} & \textbf{.122} & \textbf{.014} & \textbf{.123} & \textbf{.412} & \textbf{.060} & \textbf{.175} & \textbf{.407} & \textbf{.099} & \textbf{.167} & \textbf{.463} & \textbf{.108} & \textbf{.162} & \textbf{.361} & \textbf{.111} \\
\rowcolor{gray!8}
& \multicolumn{1}{r}{\small $\Delta (\%)$} & {\color{green!50!black} +24.9\%} & {\color{green!50!black} +9.1\%} & {\color{green!50!black} +13.0\%} & {\color{green!50!black} +22.1\%} & {\color{green!50!black} +29.6\%} & {\color{green!50!black} +25.1\%} & {\color{green!50!black} +11.3\%} & {\color{green!50!black} +57.1\%} & {\color{green!50!black} +52.7\%} & {\color{green!50!black} +49.3\%} & {\color{green!50!black} +20.1\%} & {\color{green!50!black} +20.2\%} & {\color{green!50!black} +4.1\%} & {\color{green!50!black} +12.6\%} & {\color{green!50!black} +11.8\%} & {\color{green!50!black} +44.7\%} & {\color{green!50!black} +59.5\%} & {\color{green!50!black} +48.9\%} & {\color{green!50!black} +16.6\%} & {\color{green!50!black} +14.8\%} & {\color{green!50!black} +30.4\%} \\
\midrule
\rowcolor{gray!8}
\multicolumn{23}{c}{\bf{Setting: TFs+1000}} \\
\midrule
\multirow{2}{*}{\rotatebox{90}{\small \textsc{Trad.}}}
& GENIE3 & .000 & .000 & .002 & .001 & .003 & .005 & .000 & .007 & .002 & .001 & .004 & .004 & .001 & .280 & .018 & .001 & .643 & .017 & .002 & .006 & .007 \\
& GRNBoost2 & .000 & .000 & .002 & .000 & .003 & .004 & .000 & .007 & .002 & .001 & .004 & .004 & .000 & .280 & .017 & .000 & .643 & .017 & .000 & .006 & .006 \\
\midrule
\multirow{2}{*}{\rotatebox{90}{\small \textsc{DL.}}}
& CNNC & .005 & .015 & .004 & .006 & .008 & .008 & .000 & .000 & .002 & .011 & .022 & .009 & .017 & \textit{.295} & .030 & .010 & \underline{.652} & .024 & .000 & .003 & .006 \\
& DeepSEM & .000 & .002 & .002 & .001 & .003 & .005 & .000 & .000 & .002 & .003 & .008 & .004 & .001 & .287 & .018 & .003 & .647 & .018 & .000 & .006 & .006 \\
\midrule
\multirow{4}{*}{\rotatebox{90}{\small \textsc{GNN.}}}
& GNNLink & .010 & .037 & .006 & .029 & .142 & .016 & .007 & .028 & .006 & .038 & .081 & .029 & .050 & .160 & .020 & .112 & .301 & .052 & .086 & .303 & .035 \\
& GENELink+ & .019 & .070 & .010 & .067 & \textit{.340} & .038 & \underline{.014} & \textit{.061} & \textit{.009} & \textit{.086} & \textit{.211} & \textit{.058} & \underline{.073} & \underline{.301} & \underline{.046} & \textit{.249} & \textit{.648} & \textit{.101} & \underline{.166} & \underline{.531} & \underline{.073} \\
& GRNFormer & .015 & .051 & .007 & .045 & .216 & .024 & .008 & .034 & .007 & .058 & .114 & .040 & .058 & .196 & .031 & .144 & .367 & .066 & .105 & .420 & .047 \\
& GCLink & .012 & .043 & .006 & .035 & .163 & .021 & .007 & .031 & .006 & .044 & .093 & .035 & .055 & .186 & .025 & .129 & .333 & .062 & .098 & .361 & .038 \\
\midrule
\multirow{2}{*}{\rotatebox{90}{\small \textsc{Diff.}}}
& RegDiffusion & \underline{.022} & \underline{.089} & \underline{.014} & \underline{.068} & \underline{.347} & \textit{.043} & .009 & .046 & .007 & .083 & .191 & .055 & .060 & .230 & \textit{.041} & \underline{.277} & .602 & \underline{.106} & .100 & .306 & .038 \\
& DigNet & \textit{.021} & \textit{.071} & \textit{.011} & \textit{.067} & .335 & \underline{.044} & \textit{.013} & \underline{.065} & \underline{.009} & \underline{.117} & \underline{.266} & \underline{.077} & \textit{.063} & .249 & .039 & .217 & .609 & .089 & \textit{.135} & \textit{.441} & \textit{.055} \\
\midrule
\rowcolor{gray!18}
& {\method} & \textbf{.027} & \textbf{.108} & \textbf{.015} & \textbf{.079} & \textbf{.398} & \textbf{.053} & \textbf{.016} & \textbf{.069} & \textbf{.010} & \textbf{.139} & \textbf{.299} & \textbf{.093} & \textbf{.091} & \textbf{.429} & \textbf{.051} & \textbf{.334} & \textbf{.711} & \textbf{.140} & \textbf{.276} & \textbf{.587} & \textbf{.117} \\
\rowcolor{gray!8}
& \multicolumn{1}{r}{\small $\Delta (\%)$} & {\color{green!50!black} +23.5\%} & {\color{green!50!black} +22.3\%} & {\color{green!50!black} +12.3\%} & {\color{green!50!black} +15.3\%} & {\color{green!50!black} +14.7\%} & {\color{green!50!black} +20.2\%} & {\color{green!50!black} +13.9\%} & {\color{green!50!black} +5.8\%} & {\color{green!50!black} +14.1\%} & {\color{green!50!black} +19.4\%} & {\color{green!50!black} +12.1\%} & {\color{green!50!black} +21.1\%} & {\color{green!50!black} +25.4\%} & {\color{green!50!black} +42.5\%} & {\color{green!50!black} +11.7\%} & {\color{green!50!black} +20.7\%} & {\color{green!50!black} +9.0\%} & {\color{green!50!black} +31.4\%} & {\color{green!50!black} +66.1\%} & {\color{green!50!black} +10.5\%} & {\color{green!50!black} +61.6\%} \\

\bottomrule

\end{tabular}
\end{adjustbox}
\end{spacing}
\vspace{-1mm}
\end{table}

%% file: sections/06_conclusion.tex
\section{Conclusion}    \label{sec:conclusion}
In this work, we address the gap between computational GRN inference and real-world biological discovery. We introduce {\benchmark}, an application-aligned framework that enforces an inductive, KGC-based evaluation protocol to reflect the top-$K$ constraints of wet-lab validation. To overcome the exposed gene-level generalization gap, we propose {\method}, the first joint node-edge discrete diffusion model for GRNs. By capturing the co-evolution of gene expression and network topology, and further enhanced by our TASS strategy, {\method} achieves robust inductive generalization. Together, our benchmark and model establish a unified paradigm for reliable and budget-aware biological discovery in systems biology.

\textbf{Limitations and Future Work.} 
Despite its performance, our approach has several limitations that suggest directions for future research. 
The current evaluation is restricted to established scRNA-seq systems, which may not fully represent the complexity of rare tissues or cross-species regulatory dynamics. 
We plan to extend our benchmark to include more diverse biological contexts and validate the model's transferability across different species.
Moreover, the joint diffusion process and TASS strategy introduce non-trivial computational overhead during training and inference. 
Future work will focus on developing optimized sampling kernels and parallelized subgraph processing to enhance scalability.

%% file: appendix/B_org_results.tex
\section{Results on Previous Transductive Setting}  \label{ap:org}

\input{sheets/10_a_org_specific}

\input{sheets/11_a_org_specific_auprc}

\newpage
\clearpage

%% file: sheets/10_a_org_specific.tex
\begin{table}[htbp]
\centering
\caption{Results on transductive setting (\textbf{Specific} network). Caption settings are the same as Table~\ref{tab:benchmark}.} 
\label{tab:ap:org:specific}

\setlength{\tabcolsep}{3pt}

\begin{spacing}{1.2}
\begin{adjustbox}{max width=1\linewidth}
\begin{tabular}{clccccccccccccccccccccc}

\toprule
\multicolumn{2}{c}{\multirow{2}{*}{\textbf{Model}}}
& \multicolumn{3}{c}{\bf{hESC}} 
& \multicolumn{3}{c}{\bf{hHEP}} 
& \multicolumn{3}{c}{\bf{mDC}} 
& \multicolumn{3}{c}{\bf{mESC}} 
& \multicolumn{3}{c}{\bf{mHSC-E}} 
& \multicolumn{3}{c}{\bf{mHSC-GM}} 
& \multicolumn{3}{c}{\bf{mHSC-L}} \\

\cmidrule(lr){3-5} \cmidrule(lr){6-8} \cmidrule(lr){9-11} \cmidrule(lr){12-14} 
\cmidrule(lr){15-17} \cmidrule(lr){18-20} \cmidrule(lr){21-23}

& & H@10 & H@50 & MRR & H@10 & H@50 & MRR & H@10 & H@50 & MRR 
& H@10 & H@50 & MRR & H@10 & H@50 & MRR 
& H@10 & H@50 & MRR & H@10 & H@50 & MRR \\

\midrule
\rowcolor{gray!8}
\multicolumn{23}{c}{\bf{Setting: TFs+500}} \\
\midrule
\multirow{2}{*}{\rotatebox{90}{\small \textsc{Trad.}}}
& GENIE3 & .012 & .012 & .008 & .023 & .023 & .020 & .009 & .018 & .005 & .034 & .036 & .024 & .008 & .008 & .010 & .010 & .010 & .014 & .013 & .013 & .014 \\
& GRNBoost2 & .006 & .006 & .007 & .001 & .001 & .007 & .000 & .000 & .003 & .005 & .005 & .010 & .002 & .002 & .005 & .000 & .000 & .005 & .002 & .002 & .006 \\
\midrule
\multirow{2}{*}{\rotatebox{90}{\small \textsc{DL.}}}
& CNNC & .028 & .140 & .016 & .047 & .198 & .028 & .011 & .109 & .013 & .058 & .237 & .033 & .021 & .103 & .016 & .032 & .149 & .018 & .016 & .095 & .015 \\
& DeepSEM & .024 & .071 & .012 & .054 & .059 & .030 & .045 & .073 & \textit{.022} & .094 & .149 & .042 & \textit{.078} & .086 & \textit{.049} & .074 & .075 & \textit{.053} & .065 & .067 & .036 \\
\midrule
\multirow{4}{*}{\rotatebox{90}{\small \textsc{GNN.}}}
& GNNLink & .015 & .051 & .013 & .001 & .006 & .007 & .027 & \textit{.130} & .015 & .000 & .020 & .006 & .000 & .002 & .003 & .010 & .022 & .006 & .017 & .025 & .015 \\
& GENELink+ & .040 & .169 & .022 & \textit{.137} & \textit{.377} & \textit{.057} & .016 & .071 & .010 & .103 & .358 & .053 & \underline{.329} & \underline{.598} & \underline{.167} & .048 & .213 & .029 & \underline{.130} & \underline{.330} & \underline{.057} \\
& GRNFormer & .039 & \textit{.187} & \textit{.026} & .106 & .323 & .049 & .033 & .125 & .015 & \textit{.118} & \textit{.382} & \textit{.058} & .034 & .119 & .020 & .040 & .209 & .024 & .035 & .137 & .024 \\
& GCLink & \underline{.098} & \underline{.296} & \underline{.048} & \underline{.146} & \underline{.394} & \underline{.072} & \underline{.049} & \underline{.141} & \underline{.031} & \underline{.188} & \underline{.461} & \underline{.092} & -- & -- & -- & -- & -- & -- & .058 & .213 & .032 \\
\midrule
\multirow{2}{*}{\rotatebox{90}{\small \textsc{Diff.}}}
& RegDiffusion & .031 & .148 & .017 & .035 & .140 & .019 & .022 & .060 & .018 & .053 & .176 & .025 & .040 & \textit{.208} & .024 & \textit{.087} & \textit{.311} & .044 & \textit{.081} & \textit{.257} & \textit{.036} \\
& DigNet & \textit{.043} & .168 & .023 & .055 & .198 & .029 & \textit{.049} & .120 & .019 & .064 & .281 & .031 & .049 & .184 & .025 & \underline{.132} & \underline{.434} & \underline{.066} & .048 & .235 & .028 \\
\midrule
\rowcolor{gray!18}
& {\method} & \textbf{.102} & \textbf{.318} & \textbf{.056} & \textbf{.247} & \textbf{.539} & \textbf{.128} & \textbf{.049} & \textbf{.179} & \textbf{.033} & \textbf{.259} & \textbf{.524} & \textbf{.137} & \textbf{.470} & \textbf{.769} & \textbf{.259} & \textbf{.435} & \textbf{.755} & \textbf{.227} & \textbf{.349} & \textbf{.669} & \textbf{.189} \\
\rowcolor{gray!8}
& \multicolumn{1}{r}{\small $\Delta (\%)$} & {\color{green!50!black} +4.1\%} & {\color{green!50!black} +7.3\%} & {\color{green!50!black} +15.9\%} & {\color{green!50!black} +68.7\%} & {\color{green!50!black} +36.9\%} & {\color{green!50!black} +76.3\%} & {\color{green!50!black} +0.0\%} & {\color{green!50!black} +26.9\%} & {\color{green!50!black} +4.8\%} & {\color{green!50!black} +37.6\%} & {\color{green!50!black} +13.9\%} & {\color{green!50!black} +48.7\%} & {\color{green!50!black} +42.8\%} & {\color{green!50!black} +28.7\%} & {\color{green!50!black} +54.9\%} & {\color{green!50!black} +228.7\%} & {\color{green!50!black} +74.0\%} & {\color{green!50!black} +244.1\%} & {\color{green!50!black} +167.6\%} & {\color{green!50!black} +102.7\%} & {\color{green!50!black} +231.2\%} \\
\midrule
\rowcolor{gray!8}
\multicolumn{23}{c}{\bf{Setting: TFs+1000}} \\
\midrule
\multirow{2}{*}{\rotatebox{90}{\small \textsc{Trad.}}}
& GENIE3 & .004 & .004 & .004 & .010 & .010 & .010 & .014 & .014 & .007 & .021 & .022 & .015 & .003 & .283 & .020 & .006 & .646 & .022 & .011 & .011 & .016 \\
& GRNBoost2 & .000 & .000 & .002 & .001 & .001 & .006 & .000 & .000 & .002 & .005 & .005 & .008 & .001 & .281 & .017 & .002 & .643 & .018 & .000 & .000 & .006 \\
\midrule
\multirow{2}{*}{\rotatebox{90}{\small \textsc{DL.}}}
& CNNC & .014 & .096 & .013 & .032 & .151 & .018 & .004 & .042 & .006 & .043 & .162 & .023 & .153 & .520 & .066 & .129 & .464 & .056 & .022 & .142 & .017 \\
& DeepSEM & .024 & .046 & .008 & .044 & .049 & .023 & .014 & .035 & .008 & .064 & .112 & .032 & .046 & .322 & .043 & .043 & .668 & .045 & .043 & .043 & .037 \\
\midrule
\multirow{4}{*}{\rotatebox{90}{\small \textsc{GNN.}}}
& GNNLink & .004 & .021 & .005 & .000 & .005 & .005 & .014 & .046 & .007 & .000 & .000 & .003 & .000 & .516 & .019 & .005 & .340 & .016 & .026 & .059 & .020 \\
& GENELink+ & \textit{.041} & \textit{.144} & \textit{.022} & \textit{.085} & \underline{.295} & \textit{.042} & .025 & \textit{.099} & .010 & \underline{.130} & \underline{.356} & \underline{.066} & .445 & \underline{.698} & \textit{.251} & .390 & .670 & \underline{.225} & .084 & .307 & .036 \\
& GRNFormer & .029 & .112 & .019 & .062 & .254 & .035 & .021 & .053 & .010 & .062 & .268 & .033 & \underline{.449} & \textit{.660} & \underline{.258} & \underline{.396} & \underline{.676} & \textit{.207} & .039 & .134 & .024 \\
& GCLink & \underline{.042} & \underline{.203} & \underline{.024} & \underline{.091} & \textit{.274} & \underline{.055} & \underline{.042} & \underline{.141} & \underline{.021} & \textit{.126} & \textit{.351} & \textit{.064} & -- & -- & -- & -- & -- & -- & .103 & .319 & .044 \\
\midrule
\multirow{2}{*}{\rotatebox{90}{\small \textsc{Diff.}}}
& RegDiffusion & .022 & .080 & .013 & .024 & .089 & .013 & .021 & .057 & \textit{.012} & .034 & .119 & .017 & .064 & .354 & .036 & .090 & .334 & .041 & \underline{.154} & \underline{.429} & \underline{.082} \\
& DigNet & .027 & .107 & .013 & .026 & .116 & .016 & \textit{.042} & .064 & .008 & .042 & .191 & .023 & .063 & .346 & .038 & .055 & .288 & .034 & \textit{.121} & \textit{.407} & \textit{.060} \\
\midrule
\rowcolor{gray!18}
& {\method} & \textbf{.079} & \textbf{.269} & \textbf{.035} & \textbf{.203} & \textbf{.439} & \textbf{.115} & \textbf{.053} & \textbf{.159} & \textbf{.025} & \textbf{.245} & \textbf{.502} & \textbf{.128} & \textbf{.464} & \textbf{.708} & \textbf{.267} & \textbf{.410} & \textbf{.702} & \textbf{.235} & \textbf{.311} & \textbf{.609} & \textbf{.144} \\
\rowcolor{gray!8}
& \multicolumn{1}{r}{\small $\Delta (\%)$} & {\color{green!50!black} +88.6\%} & {\color{green!50!black} +32.7\%} & {\color{green!50!black} +47.6\%} & {\color{green!50!black} +122.8\%} & {\color{green!50!black} +49.1\%} & {\color{green!50!black} +110.3\%} & {\color{green!50!black} +25.0\%} & {\color{green!50!black} +12.5\%} & {\color{green!50!black} +20.3\%} & {\color{green!50!black} +88.9\%} & {\color{green!50!black} +41.0\%} & {\color{green!50!black} +94.1\%} & {\color{green!50!black} +3.3\%} & {\color{green!50!black} +1.5\%} & {\color{green!50!black} +3.5\%} & {\color{green!50!black} +3.5\%} & {\color{green!50!black} +3.8\%} & {\color{green!50!black} +4.4\%} & {\color{green!50!black} +101.1\%} & {\color{green!50!black} +42.0\%} & {\color{green!50!black} +76.2\%} \\

\bottomrule

\end{tabular}
\end{adjustbox}
\end{spacing}
\vspace{-1mm}
\end{table}

%% file: sheets/11_a_org_specific_auprc.tex
\begin{table}[htbp]
\centering
\caption{Results of AUPRC on transductive setting (\textbf{Specific} network).}
\label{tab:ap:org:specific_auprc}

\setlength{\tabcolsep}{3pt}

\begin{spacing}{1.2}
\begin{adjustbox}{max width=1\linewidth}
\begin{tabular}{cl|ccccccc|ccccccc}
\toprule
\multicolumn{2}{c}{\multirow{2}{*}{\textbf{Model}}}
& \multicolumn{7}{c|}{\bf{TFs+500}}
& \multicolumn{7}{c}{\bf{TFs+1000}} \\

\cmidrule(lr){3-9} \cmidrule(lr){10-16}

&
& ~~hESC~~ & ~~hHEP~~ & ~~mDC~~ & ~~mESC~~ & mHSC-E & mHSC-GM & mHSC-L
& ~~hESC~~ & ~~hHEP~~ & ~~mDC~~ & ~~mESC~~ & mHSC-E & mHSC-GM & mHSC-L \\

\midrule

\multirow{2}{*}{\rotatebox{90}{\small \textsc{Trad.}}}
& GENIE3 
& .156 & .395 & .055 & .314 & .566 & .532 & .525 
& .158 & .384 & .056 & .312 & .547 & .532 & .484 \\

& GRNBoost2 
& .154 & .387 & .069 & .327 & .578 & .522 & .597 
& .150 & .380 & .059 & .324 & .547 & .527 & .481 \\

\midrule

\multirow{2}{*}{\rotatebox{90}{\small \textsc{DL.}}}
& CNNC 
& .255 & .469 & .066 & .484 & .746 & .682 & .648 
& .271 & .499 & .058 & .597 & .774 & .737 & .565 \\

& DeepSEM 
& .196 & .461 & .054 & .314 & .567 & .520 & .537 
& .192 & .419 & .056 & .320 & .565 & .543 & .523 \\

\midrule

\multirow{4}{*}{\rotatebox{90}{\small \textsc{GNN.}}}
& GNNLink 
& \underline{.521} & .751 & \textbf{.251} & .766 & .881 & .898 & .855 
& .515 & .787 & \textbf{.216} & .785 & .930 & \textit{.931} & \underline{.864} \\

& GENELink+ 
& \textit{.519} & \underline{.813} & .153 & \textit{.796} & \textbf{.943} & \textbf{.937} & \underline{.870} 
& .521 & \underline{.817} & .122 & .776 & \textbf{.954} & \textbf{.951} & .853 \\

& GRNFormer 
& .503 & .713 & .121 & \underline{.802} & .874 & .869 & .764 
& \textit{.522} & .744 & .125 & \textit{.797} & .917 & .915 & .743 \\

& GCLink 
& .465 & .707 & \underline{.207} & .739 & .856 & .849 & .814 
& .476 & .740 & \underline{.159} & .722 & .898 & .870 & .840 \\

\midrule

\multirow{2}{*}{\rotatebox{90}{\small \textsc{Diff.}}}
& RegDiffusion 
& .447 & \textit{.808} & .117 & .761 & \underline{.932} & \underline{.925} & \textit{.863} 
& .483 & \textit{.814} & .132 & .784 & \underline{.944} & \underline{.944} & \textit{.859} \\

& DigNet 
& .495 & .799 & .112 & .775 & .717 & .703 & \textbf{.871} 
& \underline{.527} & .813 & .139 & \underline{.798} & .625 & .612 & \textbf{.869} \\

\midrule

\rowcolor{gray!18}

& {\method} 
& \textbf{.582} & \textbf{.817} & \textit{.155} & \textbf{.836} & \textit{.923} & \textit{.909} & .840 
& \textbf{.598} & \textbf{.822} & \textit{.154} & \textbf{.856} & \textit{.931} & .926 & .839 \\

\rowcolor{gray!8}

& \multicolumn{1}{r}{\small $\Delta (\%)$} 
& {\color{green!50!black} +11.8\%} & {\color{green!50!black} +0.5\%} & {\color{red!70!black} -38.4\%} & {\color{green!50!black} +4.3\%} & {\color{red!70!black} -2.0\%} & {\color{red!70!black} -3.0\%} & {\color{red!70!black} -3.6\%} 
& {\color{green!50!black} +13.3\%} & {\color{green!50!black} +0.5\%} & {\color{red!70!black} -28.4\%} & {\color{green!50!black} +7.2\%} & {\color{red!70!black} -2.4\%} & {\color{red!70!black} -2.6\%} & {\color{red!70!black} -3.5\%} \\

\bottomrule
\end{tabular}
\end{adjustbox}
\end{spacing}
\vspace{-1mm}
\end{table}

%% file: appendix/E_cell_discrete.tex
\section{Cell-cluster Discretization}   \label{ap:cell}

This section provides additional details and supporting evidence for the proposed Cell-cluster-based Discretization strategy. We first present the algorithmic procedure, followed by visualizations that demonstrate the biological coherence of the induced discrete gene state space.

\subsection{Algorithm}    \label{ap:cell:alg}

\begin{algorithm}[htbp]
\caption{Cell-cluster Discretization}
\label{alg:discretization}
\begin{algorithmic}[1]
\Require Gene expression matrix $\mathbf{X} \in \mathbb{R}^{N \times F_X}$, number of clusters $k$
\Ensure Discrete one-hot representations $\mathbf{D} \in \mathbb{R}^{N \times F_D}$, where $F_D = 2^k$
\State Construct $\mathbf{X_{vis}}$ using only visible data
\State Apply $k$-means on cell-wise expression profiles of $\mathbf{X_{vis}}$ to obtain clusters $\{\mathcal{C}_1, \dots, \mathcal{C}_k\}$
\State For each cluster $\mathcal{C}_j$, compute $
\tau_j = \frac{1}{|\mathcal{C}_j| \cdot N} \sum_{c \in \mathcal{C}_j} \sum_{g=1}^{N} \mathbf{X}_{g,c}
$
\For{each gene $g$}
    \For{each cluster $j$}
        \State Compute mean expression $\mu_{g,j} = \frac{1}{|\mathcal{C}_j|} \sum_{c \in \mathcal{C}_j} \mathbf{X}_{g,c}$
        \State Set binary activation $b_{g,j} = \mathbb{I}(\mu_{g,j} \ge \tau_j)$
    \EndFor
    \State Form binary vector $\mathbf{b}_g \in \{0,1\}^k$
    \State Map $\mathbf{b}_g$ to category index $c_g \in \{0, \dots, F_D - 1\}$
    \State Convert $c_g$ to one-hot vector $\mathbf{d} \in \mathbb{R}^{F_D}$
\EndFor
\State \textbf{return} $\mathbf{D} = \{\mathbf{d}\}_{g=1}^{N}$
\end{algorithmic}
\end{algorithm}

\subsection{Design Motivation and Biological Evidence} \label{ap:cell:vis}

We provide the design motivation and biological evidence supporting the proposed Cell-cluster-based Discretization. The goal of this design is to construct a biologically coherent discrete state space that enables stable inductive diffusion over unseen genes.

\begin{figure}[htbp]
\centering
    \includegraphics[width=1.0\linewidth]{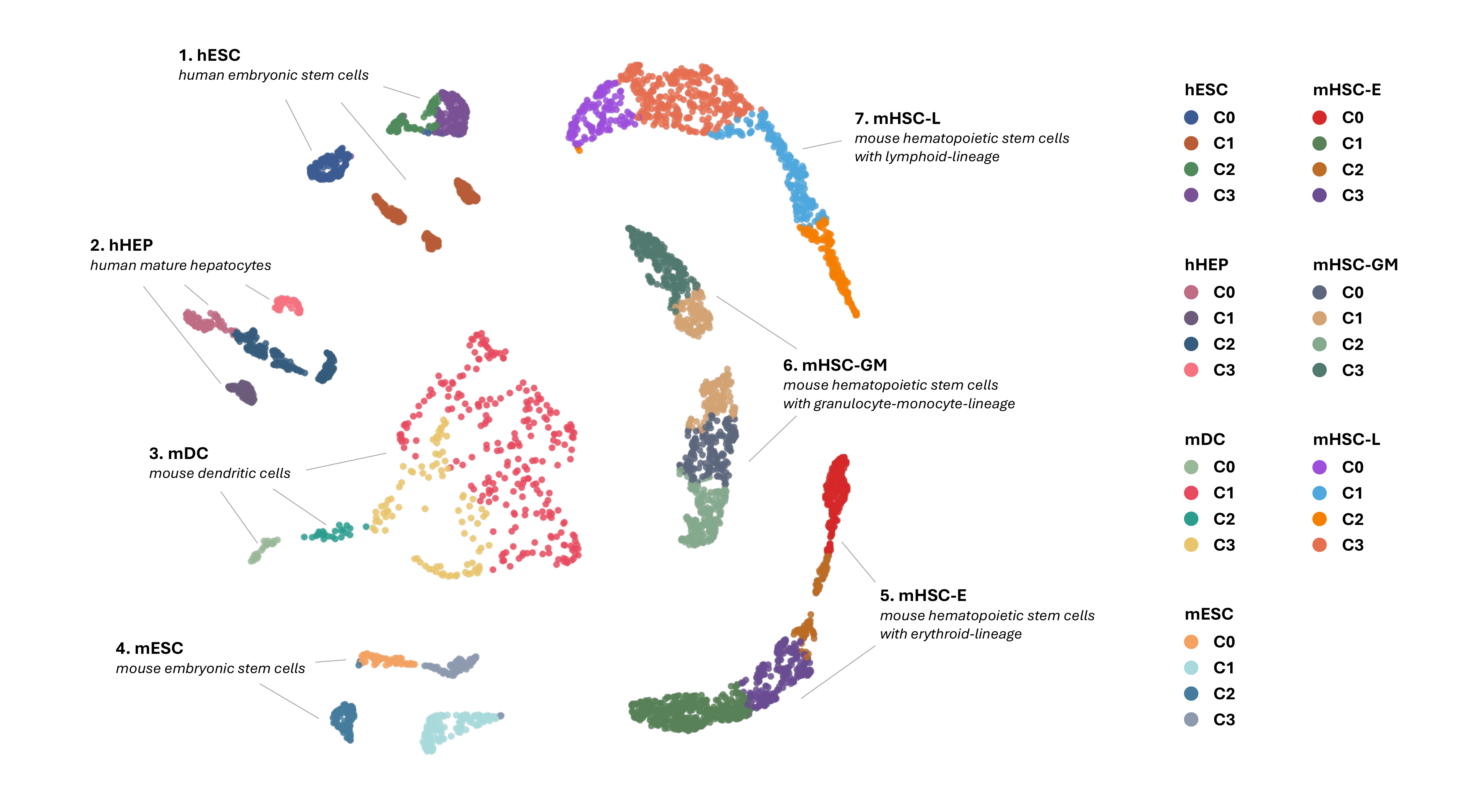}
    \vspace{-5.0mm}
    \caption{\textbf{Cell-state structure induced by clustering.} 
    Visualization of cell clusters obtained from scRNA-seq profiles using UMAP~\citep{becht2019dimensionality}. Each cluster corresponds to a distinct cellular state, supporting the use of cell-level structure as the basis for discretization.}
    \label{fig:cell_cluster}
\end{figure}

\textbf{Motivation.} 
A natural alternative is to directly cluster gene expression profiles using standard unsupervised methods (e.g., k-means). However, such approaches may produce clusters dominated by statistical artifacts such as sparsity patterns or zero inflation in scRNA-seq data, rather than biologically meaningful structure. In contrast, scRNA-seq measurements are inherently organized by cellular contexts (e.g., conditions or states), suggesting that discretization should be grounded in cell-level structure to preserve functional coherence.

\begin{figure}[htbp]
\centering
    \includegraphics[width=1.0\linewidth]{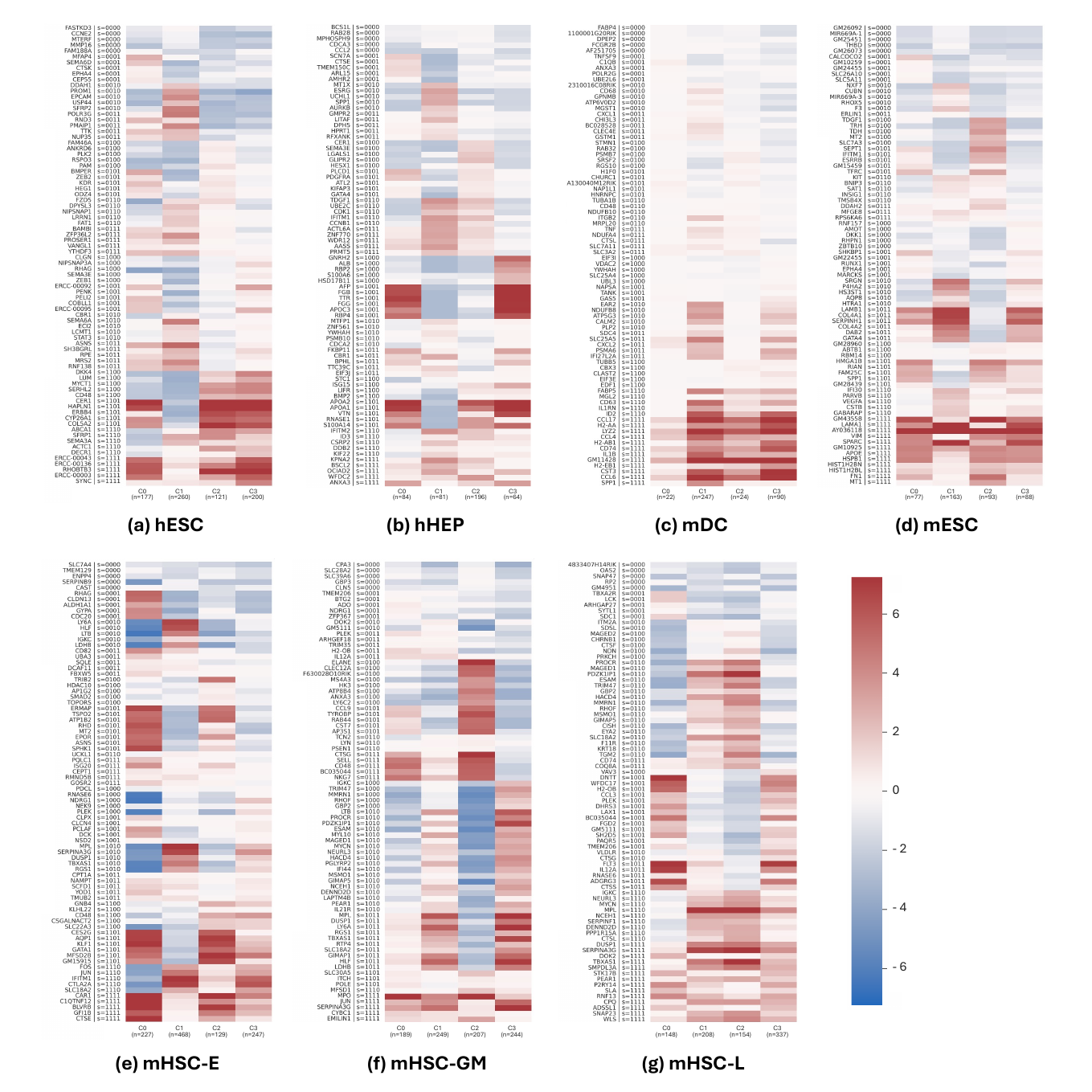}
    \caption{\textbf{Gene activation consistency across cell clusters.} 
    Representative genes show stable and coherent activation patterns within each cell cluster, indicating that discretization preserves biologically meaningful gene activity structure.}
    \label{fig:repre_gene}
\end{figure}

\textbf{Design Principle.} 
Our method explicitly leverages this structure by first clustering cells into $k$ biologically meaningful groups, and then constructing gene-level discrete representations based on expression consistency across these cell states. This ensures that discretization reflects stable functional activity rather than superficial similarity in gene expression vectors.

\textbf{Figure Organization and Evidence.} 
We use UMAP~\citep{becht2019dimensionality} to visualize the resulting cell clusters obtained from scRNA-seq data in Figure~\ref{fig:cell_cluster}. The clusters correspond to distinct cellular states, indicating that the clustering captures meaningful biological organization rather than arbitrary partitions.

\begin{figure}[htbp]
\centering
    \includegraphics[width=1.0\linewidth]{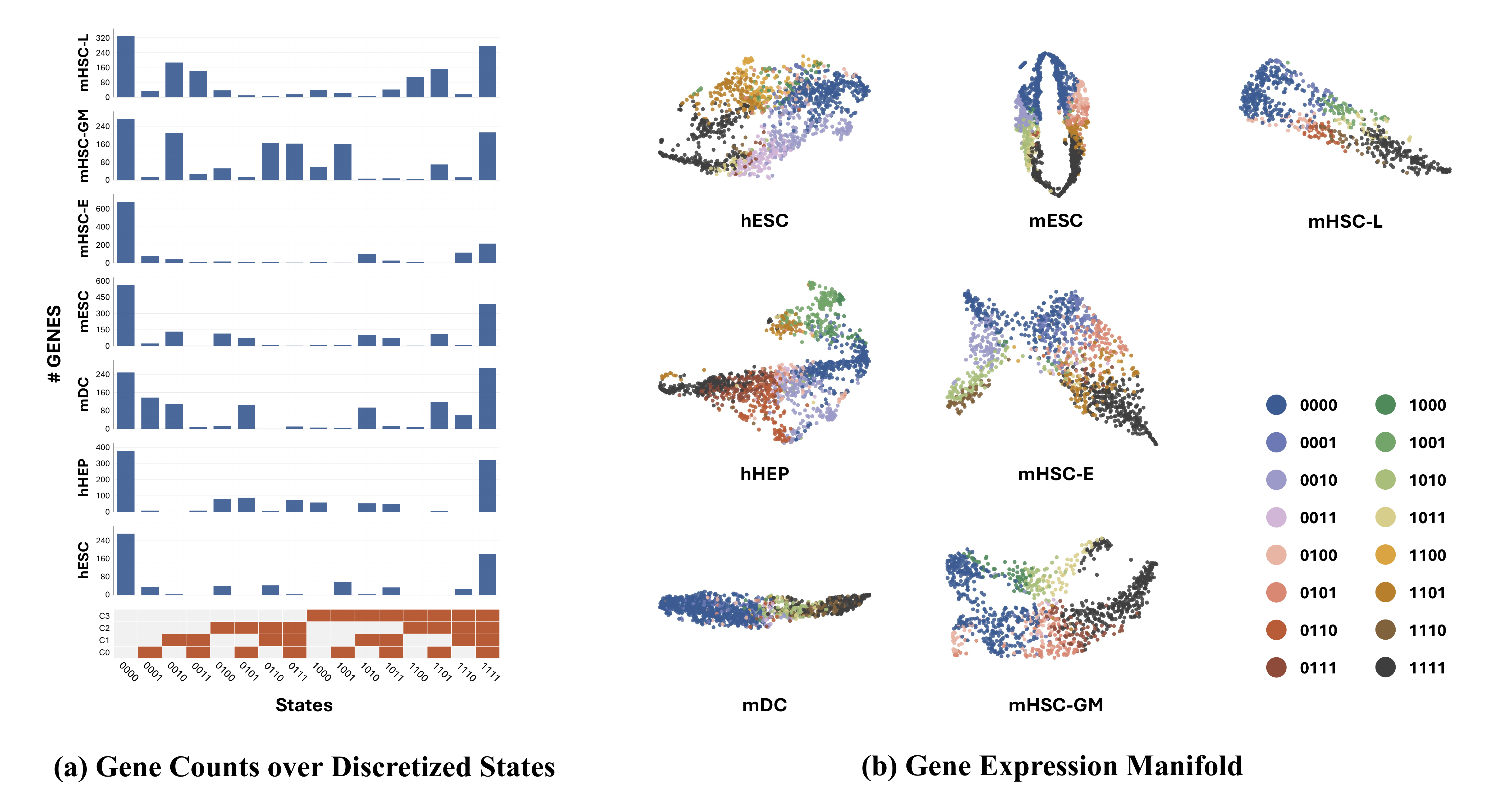}
    \vspace{-5.0mm}
    \caption{\textbf{Discrete gene manifold structure.} 
    The induced gene representation space exhibits smooth and coherent structure, indicating that the discretization preserves biologically meaningful organization while enabling stable diffusion over discrete states.}
    \label{fig:gene_dist}
\end{figure}

Figure~\ref{fig:repre_gene} presents representative genes across different cell clusters. We observe that genes exhibit consistent activation patterns within clusters, suggesting that discretization captures stable gene-level functional behavior conditioned on cellular context. This is further proved by Figure~\ref{fig:gene_dist}, which visualizes the induced gene representation manifold in the discrete state space using UMAP~\citep{becht2019dimensionality}. The resulting structure exhibits smooth organization and local coherence, suggesting that the discretization preserves biologically meaningful continuity while providing a structured state space suitable for diffusion modeling.

Together, these results support the role of Cell-cluster-based Discretization as a biologically grounded representation scheme that enables stable inductive generalization under sparse and heterogeneous scRNA-seq data.

\newpage

%% file: appendix/F_tass.tex
\section{TF-ALL Subgraph Sampling (TASS)}    \label{ap:tass}

In this section, we present the complete pipeline of TF-ALL Subgraph Sampling (TASS), which consists of a stochastic subgraph sampling strategy followed by a consensus-based reconstruction procedure for global edge inference.

\subsection{Sampling}  \label{ap:tass:sampling}

\begin{algorithm}[htbp]
\caption{TF-ALL Subgraph Sampling (TASS)}
\label{alg:tass_sampling}
\begin{algorithmic}[1]
\Require Graph $G=(V,E)$, TF set $V_{\text{TF}}$, gene set $V_{\text{G}}$, subgraph size $k$
\Ensure Subgraph collection $\mathcal{S} = \{G_1, \dots, G_m\}$

\State Compute $n \leftarrow |V|$, $p \leftarrow k / n$, TF-ALL ratio $t \gets |V_{\text{TF}}| / |V_{\text{G}}|$
\State Calculate theoretical bound $m_{base} \leftarrow \left\lceil p^{-2} \log n \log(1/\delta) \right\rceil$
\State Set $m = 100 \cdot \lceil m_{base} / 100 \rceil$ \Comment{Round up to nearest multiple of 100}
\State Initialize $\mathcal{S} \leftarrow \emptyset$

\For{$i = 1$ to $m$}
    \State Sample TF nodes: $V^{(i)}_{\text{TF}} \sim \text{Uniform}(V_{\text{TF}}, \lfloor |V_{\text{TF}}| \cdot t \rfloor)$
    \State Sample gene nodes: $V^{(i)}_{\text{G}} \sim \text{Uniform}(V_{\text{G}}, k - |V^{(i)}_{\text{TF}}|)$
    \State Duplicates $V_{\text{dup}} \gets V^{(i)}_{\text{TF}} \cap V^{(i)}_{\text{G}}$, $n_{\text{dup}} \gets |V_{\text{dup}}|$
    \While{$n_{\text{dup}} \ne 0$}  \Comment{Resolve duplicates}
        \State Resample a new gene $g \sim \text{Uniform}(V_\text{G})$
        \If{$g \not \in V^{(i)}_{\text{TF}} \cup V^{(i)}_{\text{G}}$}
            \State Add $g$ into the $V^{(i)}_{\text{G}}$, $n_{\text{dup}} \gets n_{\text{dup}} -1$
        \EndIf
    \EndWhile
    \State Construct induced subgraph: $G_i = G[V^{(i)}_{\text{TF}} \cup V^{(i)}_{\text{G}}]$
    \State $\mathcal{S} \leftarrow \mathcal{S} \cup \{G_i\}$
\EndFor

\State \textbf{return} $\mathcal{S}$
\end{algorithmic}
\end{algorithm}

\subsection{Reconstruction}  \label{ap:tass:reconstruct}

\begin{algorithm}[htbp]
\caption{TF-ALL Subgraph Sampling (TASS): Reconstruction}
\label{alg:tass_reconstruction}
\begin{algorithmic}[1]
\Require Global graph $G=(V,E)$, sampled node sets $\mathcal{V} = \{V_i\}_{i=1}^{m}$, trained model $\Phi$, placeholder $\lambda$
\Ensure Global edge score matrix $\mathbf{S} \in \mathbb{R}^{|V|\times|V|}$

\State Initialize accumulation matrix $\mathbf{A} \leftarrow \mathbf{0}$, count matrix $\mathbf{C} \leftarrow \mathbf{0}$

\For{each sampled node set $V_i \in \mathcal{V}$}
    \State Construct induced subgraph $G_i = G[V_i]$
    
    \State Predict raw edge logits: $\mathbf{E}_i = \Phi(G_i)$
    
    \State Normalize edge scores: $\mathbf{P}_i = \text{Softmax}(\mathbf{E}_i)$ \Comment{within-subgraph normalization}
    
    \State Map local edges to global index space: $(U,V) \leftarrow \pi(V_i \times V_i)$
    
    \State Accumulate predictions: $\mathbf{A}_{U,V} \leftarrow \mathbf{A}_{U,V} + \mathbf{P}_i, \mathbf{C}_{U,V} \leftarrow \mathbf{C}_{U,V} + 1$
\EndFor

\State \textbf{Consensus aggregation:}
\For{each potential edge $(u,v) \in V \times V$}
    \If{$\mathbf{C}_{u,v} > 0$}
        \State $\mathbf{S}_{u,v} = \mathbf{A}_{u,v} / \mathbf{C}_{u,v}$ \Comment{mean across subgraphs}
    \Else
        \State $\mathbf{S}_{u,v} = \lambda$ \Comment{unobserved edge prior}
    \EndIf
\EndFor

\State \textbf{return} $\mathbf{S}$
\end{algorithmic}
\end{algorithm}

\newpage

%% file: appendix/G_BEELINE.tex
\section{Dataset Information of {\benchmark}}  \label{ap:bench}

The original BEELINE benchmark is constructed from five experimental single-cell RNA-seq datasets covering seven cell types (hESC, hHEP, mDC, mESC, mHSC-E, mHSC-GM, mHSC-L) across human and mouse systems~\citep{pratapa2020benchmarking}. These datasets span diverse biological contexts, including differentiation and developmental processes, and provide measurements that enable pseudotime inference.

Each dataset is paired with up to four types of ground-truth networks:: cell-type-specific (Specific) network derived from ChIP-seq experiments and curated databases such as ENCODE~\citep{kagda2025data}, ChIP-Atlas~\citep{zou2024chip}, and ESCAPE~\citep{xu2013escape}; 
non-specific (Non-Specific) network aggregated from large-scale resources such as DoRothEA~\citep{garcia2019benchmark}, RegNetwork~\citep{liu2015regnetwork}, and TRRUST~\citep{han2018trrust}; 
functional interaction networks from STRING~\citep{szklarczyk2023string}, capturing broader biological associations that may include indirect regulatory relationships; 
loss-/gain-of-function (LOF/GOF) evidence from ESCAPE~\citep{xu2013escape} (available only for mESC).
For each dataset, two gene subsets are constructed by selecting the top 500 or 1{,}000 most variably expressed genes with transcription factors (\emph{Abbr.}: TFs+500 and TFs+1000, respectively), culminating in total 44 distinct evaluation tasks.

Among these settings, future work may place particular emphasis on the Specific networks, as their cell-type-matched regulatory evidence provides the closest alignment between the expression measurements and the evaluation targets, making them a more representative testbed for context-dependent GRN discovery. The other network types remain valuable as complementary settings for assessing robustness to broader, heterogeneous, or functional interaction evidence.